# Decision Methods for Adaptive Task-Sharing in Associate Systems

**Thomas S. Paterson**
Laboratory for Intelligent Systems
Dept. of Engineering-Economic Systems
Stanford University
Stanford, California 94305

**Michael R. Fehling**
Laboratory for Intelligent Systems
Dept. of Engineering-Economic Systems
Stanford University
Stanford, California 94305

## Abstract

This paper describes some results of research on associate systems: knowledge-based systems that flexibly and adaptively support their human users in carrying out complex, time-dependent problem-solving tasks under uncertainty. Based on principles derived from decision theory and decision analysis, a problem-solving approach is presented which can overcome many of the limitations of traditional expert-systems. This approach implements an explicit model of the human user's problem-solving capabilities as an integral element in the overall problem solving architecture. This integrated model, represented as an influence diagram, is the basis for achieving adaptive task sharing behavior between the associate system and the human user. This associate system model has been applied toward ongoing research on a Mars Rover Manager's Associate (MRMA). MRMA's role would be to manage a small fleet of robotic rovers on the Martian surface. The paper describes results for a specific scenario where MRMA examines the benefits and costs of consulting human experts on Earth to assist a Mars rover with a complex resource management decision.

## 1 Introduction

This paper describes some results of research on associate systems: knowledge-based systems that manage their own task performance to flexibly and adaptively support their human users in carrying out complex, time-dependent problem-solving tasks under uncertainty. Our efforts to develop such systems focus on the development of problem-solving methods based on well founded principles of time-bounded rationality [Fehling & Breese, 1990; Horvitz, 1988; Russell & Wefald, 1991]. The principles which underlie the approach we shall describe are derived from decision theory and decision analysis [Raiffa, 1968; Howard & Matheson, 1984]. We believe that this principled approach to problem solving can overcome many of the limitations of traditional expert-system approaches, particularly in applications where the system must flexibly and cooperatively share task-performance with human users.

Conventional expert system approaches suffer from some critical shortcomings in providing adaptive and cooperative support to their users. For the most part, conventional knowledge systems, especially so-called rule-based systems, support their users with problem-solving expertise that is encoded in the form of fixed problem representations and solution methods. The support provided by such systems breaks down in situations that are not sufficiently represented, or when the solution methods do not match the style or perspective of the user. In addition, knowledge systems are designed with a fixed commitment to the role played - i.e., the tasks performed - by the expert system; there is no ability to cooperatively adapt the actions of the support system to the situation-specific needs of its user. Furthermore, conventional systems suffer from an inability to adapt to the concerns of a user to the extent that these concerns diverge from those of the expert whose knowledge is encapsulated.

The source of these shortcomings in conventional knowledge systems is the failure to recognize the context-sensitive needs and objectives of the human user. The user may require any range of support, from an analytical tool to a domain tutor, but it is the user's preferences which should drive the system's actions in providing that support. We refer to this type of system as an **associate**: one which acts flexibly, adaptively, and cooperatively in support of human problem-solving in performing complex, time-dependent tasks under uncertainty.

Our focus in this paper is on the management of task-sharing between the human user and an associate system. We will illustrate how a decision analytic framework for problem solving in an associate system provides the mechanism for adaptive task sharing under uncertainty. Preliminary results are presented for an associate system application dealing with the management of robotic planetary exploration rovers.

## 2  Associate Systems

Associate systems bridge the gap between systems that perform tasks "on-line", in real-time, and with complete autonomy in initiating and completing tasks, and typical advisory systems that operate "off-line" in giving advice to humans who must initiate and complete all primary task activities.  Advances in associate system architectures were recently addressed at a DARPA sponsored conference [Lehner, P., 1991]. Two notable associate system programs discussed at the conference are the U.S. Air Force's Pilot's Associate (PA), and the U.S. Navy's Submarine Operation Automation System (SOAS).

As in any cooperative relationship, both the associate system and the human user make use of their perceptions of each other's relative strengths and weaknesses. While humans are good at making judgements from synthesising large amounts of information, they have a limited capacity and efficiency for consistently applying principles for correct problem solving.  Computation systems, including intelligent systems, are good at following such normative principles, but their capacity for judgement is limited to their internal knowledge representations and inference capabilities.  Associate systems exploit these relative strengths by cooperatively focusing on performance of subtasks involving minimal judgement.

In addition, associate systems exhibit context sensitive *mixed-initiative* behavior by adapting the range of subtasks they perform to meet the demands of the current problem solving context.  For example, an associate system may weight the costs (e.g. time) versus the benefits (e.g. validity) of consulting the human user for his/her judgement before taking action autonomously. Alternatives to consult the user or not are evaluated according to a utility function which expresses the user's preferences over multiple, possibly competing, task objectives.  Since the decision is made under uncertainty, maximum expected utility is the criterion for selecting a decision alternative.  This utility-maximizing approach provides the mechanism for evaluating the costs and benefits of consulting the human user.

As the preceding example suggests, the key to an associate system's mixed-initiative behavior is the adequacy of the consultation decision. This decision must reflect the potential benefits of attaining the human user's judgement before completing and executing a plan for some specific course of one or more actions. Consultation might improve the state of knowledge of the associate to support planning and execution, increasing the expected utility of the problem solving actions that are likely to be undertaken. The decision must also reflect the costs of making the consultation with the human user. These costs should reflect not only resource or performance costs from the associate's point of view, but also the costs incurred by the human user as well.

Uncertainties and situational dependancies of both benefits and costs must be modeled. The outcome of task-performance and, the cost of taking those actions, including the cost (if any) of making the consultation, all effect the final outcome achieved by the cooperative human-associate problem solving system. The estimated state of this *total system* is used to determine the comparative utility of the various strategies for task sharing.

The primary model, then, for an associate system is an integrated dual decision problem:  whether or not to consult the human user prior to taking action, and what action to take. We represent this model in an influence diagram [Howard & Matheson, 1984], as shown in Figure 1.

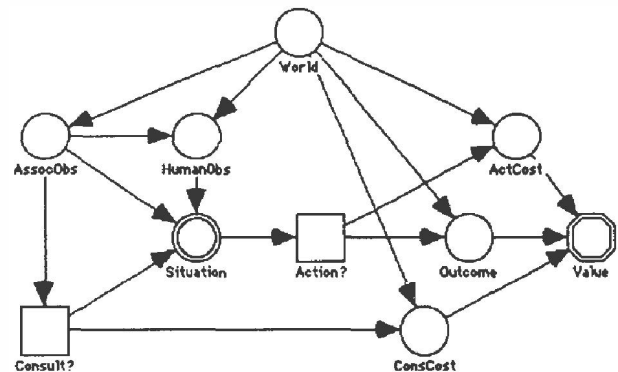

Figure 1:  Generic Associate System Influence Diagram

Round nodes in the influence diagram represent state variables whose value is uncertain.  Two-circle nodes represent deterministic functions whose output is purely a function of the inputs with no additional uncertainty. Square nodes represent decision variables, and octagons represent the utility function which expresses the preferences of the human user. Arrows between uncertain nodes represent probabilistic dependencies between the two variables.  Arrows pointing to decision nodes indicate that the state of the node at the tail of the arrow is known when the decision is made. While a node in an influence diagram is intended to represent a single state variable, each node in Figure 1 may be seen as a "super-node" representing a vector of many state variables.

The **World** node in Figure 1 represents the uncertain state of the task environment with which the human-associate system is interacting. The associate is capable of making imperfect observations of the world (**AssocObs**) which it can then use to determine the **Situation** with which it is dealing. Those observations are known at the time that it makes the decision whether or not to **Consult** the human user. Depending on the implementation, the human may also be capable of making direct, but imperfect, observations of the world (**HumanObs**), as well as having access to the observations made by the associate.

If consulted, the human observations are also used to determine the situation prior to making a decision on what **Action** to take. The **Situation** is modeled as a deterministic node, since it represents a deterministic procedure by which the associate will synthesise both the associate and human observations of the world. Both the action itself and the state of the world influence the **Outcome** of the action, as well as the costs incurred by taking that action (**ActCost**). If the human user is consulted there will be some consultation costs (**ConsCost**) which could also be influenced by the state of the world. The **Value** of the consult decision together with the action decision is a function of the outcome of the action and total costs incurred during consultation with the human user and the course of action carried out.

The consultation model discussed above is made up of the **HumanObs** and **ConsCost** nodes. It is the **HumanObs** node which captures the consulting benefits or judgement capabilities of the human user. Characterizing the possible states and corresponding conditional probabilities for this node is a significant knowledge acquisition task which should not be underestimated.

## 3 The Mars Rover Manager's Associate (MRMA) system

We have designed and implemented a system that demonstrates management of mixed-initiative behavior in a Mars Rover Manager's Associate (MRMA). Intended as part of an unmanned robotic exploratory mission to Mars, MRMA's role would be to manage a small fleet of semi-autonomous rovers on the Martian surface. MRMA itself would be located on a Mars-orbiting communications satellite serving as the link between the rovers and the human Rover Manager (RM) at mission control on Earth.

The management of Mars rovers is an excellent application for a decision making associate system such as MRMA. Uncertainties will have serious impact on virtually every aspect of the mission: environment, communications, navigation, rover performance, etc. Important tradeoffs will have to be made between rover safety (i.e. rover mission lifetime) and the potential gain of scientific knowledge.

While the rovers themselves could be designed to exhibit associate behavior, there are several advantages to having a centralized intelligent management system. Environmental knowledge bases would be maintained with MRMA so that each rover could learn from the experiences of others in the fleet. Integrated planning solutions for cooperating rovers could also be formulated by MRMA. While strategic and tactical planning would be executed by MRMA, only lower level operational planning would be required of the rovers themselves. By offloading these functions from the rovers, they would be smaller and less expensive, with more of them for a given mass budget for Earth-to-Mars transport. Thus, an associate management system would complement the use of small, intelligent "insect" robots, such as those developed by Rodney Brooks at MIT.[1]

Resource management will be a dominant theme in managing the rover fleet. Various consumables will be carried on board to carry out scientific experiments. Sample collection mechanisms as well as the propulsion system will exhibit significant wear and tear over the mission lifetime. While the rovers will not have to manage battery energy as a resource, they will have to manage time. Current designs for the rovers make use of radioisotope thermal generators or RTGs as their power supplies. Once power generation is initiated in the RTGs they cannot be powered down during periods of inactivity; the power would be lost, radiated as heat. Therefore, it is critical that inactivity periods for the rovers are minimized to make use of their limited mission lifetime (2 to 5 years for the RTGs). Due to the long time delay for Mars-Earth communications (10 to 45 minutes round trip, varying with a period of about 2 years), there is a high cost associated with consulting mission control for human judgement. This cost is even higher (and more uncertain) when the time for human analysis of the returned data is considered.

This time delay constitutes the cost component of the mission control consultation model. Capturing accurate expectations of human judgement capabilities for the consultation model will require characterizing the performance of mission control personnel in interpreting rover sensor data. One obvious example of this is visual scene interpretation. Although the rover will posess image processing and recognition software, the performance of current technology in this area is limited. Knowledge acquisition tests with the human rover managers would be performed to assemble conditional probabilities for properly identifying various classes of surface features. Similar performance characterizations would be made of the rover hardware/software as well. Both rover and human performance knowledge bases would be updated during the course of the mission as both gain experience.

We now present an example where MRMA must decide whether or not to consult the human rover manager before planning a deviation from a rover's nominal path. The scenario is illustrated in Figure 2.

---

[1] MRMA is, in fact, part of an overall demonstration that includes these components. Work on a case-based planning system and the actual robotic devices is being performed by ISX Corporation.

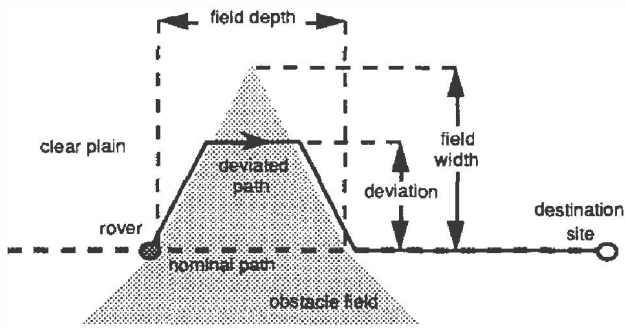

Figure 2: Rover Path Deviation Scenario

Along its nominal path, the rover has encountered an obstacle field whose surface characteristics differ from the clear plain over which it had been traveling. Due to limited resolution in mapping the Martian surface by satellite, the depth and width of the field are uncertain. Prior estimates from satellite maps indicate the field is strewn with rocks of uncertain size. Depending on the size of the rocks, the rover may be able to roll or step over them with little or no change in speed, or it may have to reduce its speed significantly to move around them. To minimize the time required for the rover to reach its next destination site, MRMA must decide how far the rover should deviate around the field. While MRMA must use prior estimates on the fields dimensions from the satellite maps, it can task the rover to process a far-field visual scene of the obstacle field (since the rover is currently at the edge). The processed image provides an improved estimate of rock size. However, MRMA also has the option of transmitting the visual scene to the human rover managers at mission control for them to interpret. Figure 3 shows how we have instantiated the generic associate system influence diagram (Figure 1) for this scenario.

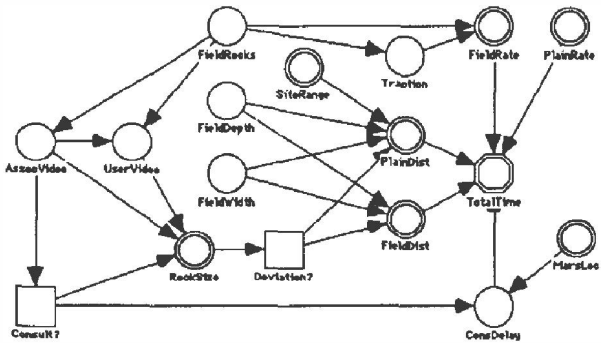

Figure 3: Associate System Influence Diagram for Path Deviation Scenario

The deviated path is broken into two segments: that which lies in the plain, and that which lies in the field. The distances of both segments (**PlainDist** and **FieldDist**) are functions of the uncertain field dimensions (**FieldDepth** and **FieldWidth**), the path **Deviation**, and the range from the rover's present location to the destination site (**SiteRange**). The **FieldRocks** can be interpreted by both the associate and the rover manager (**AssocVideo** and **UserVideo**) to derive an estimate of the **RockSize**. The rover's rate through the field (**FieldRate**) is a function of the **FieldRocks** and **Traction**, while the rate over the plain (**PlainRate**) is assumed to be known. The location of Mars in its orbit relative to the Earth (**MarsLoc**) is relevant to the consultation delay (**ConsDelay**) with the rover manager. In this example, the utility function is one dimensional, being concerned only with the **TotalTime** required for the rover to get to the destination site.

It is obvious from our description that a significant level of knowledge representation is required to assess and implement the associate system decision model. The conditional probability distributions for the uncertain nodes capture prior knowledge about the environment (e.g. **FieldRocks**), knowledge of sensor accuracies (e.g. **FieldDepth, FieldWidth**), and performance knowledge of the associate and human user (e.g. **AssocVideo, Traction, UserVideo, ConsDelay**). Other performance knowledge is captured in the deterministic node functions (e.g. **FieldRate, PlainRate**).

While many efficient algorithms exist for solving influence diagrams [Olmsted, 1983; Shachter, 1986], various types of analyses can be performed on an influence diagram prior to actual solution. Deterministic sensitivity analysis and value-of-information calculations are of particular use to associate systems [Howard, Matheson 1984]. Deterministic sensitivity analysis is used to evaluate the sensitivity of the value node's utility function to the various sources of uncertainty in the influence diagram. For each combination of decision alternatives, the utility range is determined for each variable by fixing the remaining variables at their base case values. Ranking each variable in its contribution to total variance, the utility ranges for each variable can be plotted in a "tornado diagram". Figure 4 shows one such tornado diagram for the influence diagram in Figure 3.

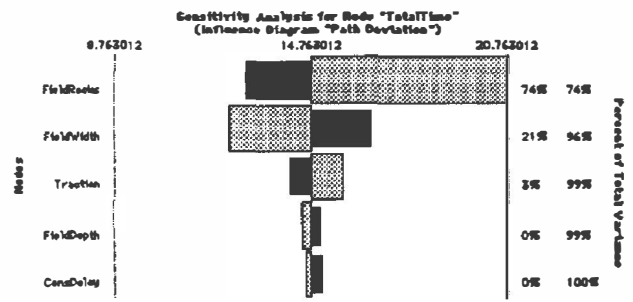

Figure 4: Tornado Diagram for Path Deviation Scenario

The associate then has the capability to fix uncertain variables at their base case values if they fall below a predetermined cumulative variance threshold (e.g. 95%). Since fixing uncertain variables reduces the combinatorial complexity of the problem, this analysis aids the associate in achieving time-bounded rationality. If the human user is consulted for advice, or if the user is reviewing actions that the associate has taken autonomously, the tornado diagram is a useful representation for the user to gain insight onto the problem itself, and the problem solving approach of the associate.

For the analysis shown in Figure 4, the **FieldRocks** node was shown to be the most sensitive of all the uncertainties in the model. This is fortunate, since the **AssocVideo** and **UserVideo** nodes provide (imperfect) information on the state of **FieldRocks** prior to making the **Deviation** decision. Figure 4 also shows a high sensitivity to **FieldWidth**. Value-of-information calculations could be performed to determine the cost (in time) that the associate/user would be willing to pay for perfect information on this variable (in terms of the influence diagram, the increase in utility from adding an information arrow from **FieldWidth** to **Deviation**). For this particular scenario, the associate might expand the influence diagram shown in Figure 3 to include a third decision node to examine the alternative of waiting for the next mapping satellite pass for an improved estimate on the distribution of **FieldWidth**.

We have solved the decision model in Figure 3 using a number of numerical assessments. The optimal policies which MRMA developed for the **Consult** and **Deviation** decisions show examples of both autonomous and consulting behavior, according to its prior beliefs about its environment, MRMA's own performance at interpreting its environment, as well as the performance of the human user.

MRMA is currently implemented both in HyperCard on a Macintosh II, and using CLOS in Macintosh Common Lisp (MCL). While the HyperCard implementation uses conventional discrete probability distributions and discrete decision alternatives, the MCL implementation uses a linear-quadratic-Gaussian model with multivariate normal distributions [Shachter & Kenley, 1989].

## 4 Analysis and Review

Our research with MRMA is helping us to formulate and implement problem solving capabilities required in associate systems and to demonstrate their value in performing complex time-bounded tasks under uncertainty. While other research efforts have shown the effectiveness of decision theoretic approaches for adaptive, automated problem-solving under uncertainty [Fehling & Breese, 1990; Horvitz, 1988; Russell & Wefald, 1991], our efforts on MRMA have clarifiied several issues in the management of task sharing in associate systems. Specifically, we have developed the model of the consultation decision which the associate uses to infer the utility of deferring its own action in order to consult the human user. Our work has also expanded the notion that intelligent associate systems should employ decision theoretic computations to guide problem solving by using such techniques to determine appropriate context-specific tactics for person-machine, cooperative task sharing.

This research has shown that the methods of rational decision making are central to associate system design. While these decision methods constitute a normative model of problem-solving, we believe the concept and implementation of the consultation model for associates seeking support from humans is very similar to the process humans use when contemplating the use of software tools in time or resource constrained situations. Finally, our work with MRMA has led to the design and application of new software tools and architectures for implementing decision theoretic principles in associate systems.

## 5 Implications and Future Work

Work on associate systems extends the use of decision theoretic methods to an important point in the development of intelligent system technology, enabling cooperative person-machine interaction. Decision theoretic methods are critically needed to provide a principled basis for reasoning and acting under uncertainty. Moreover, these decision principles focus system design and operation on serving a user's preferences rather than meeting a fixed set of goals that may or may not arise out of these preferences depending upon context. We are continuing to explore these issues with further development of MRMA and with associate systems for other problem solving applications such as intelligent management of complex, distributed production processes, and in the development of an instructable robot.

We are exploring new knowledge representation techniques that are better suited for use in an associate system's decision making and action control processes. These knowledge representation techniques support uniform representation of uncertain causal, taxonomic, and temporal relations among events, objects, and process concepts. We are also developing methods for dynamic model construction [Fehling & Johnson, 1991] that allow an associate system to automate construction of an influence diagram of the form built by hand in MRMA. This work builds upon earlier research on automated decision modeling [Holtzman, 1989].